\documentclass[letterpaper, 10 pt, conference]{ieeeconf}

\IEEEoverridecommandlockouts
\usepackage{cite}
\usepackage{url}
\usepackage{amsmath,amssymb,amsfonts}
\usepackage{graphicx}
\usepackage{textcomp}
\usepackage{xcolor}

\usepackage{booktabs}
\usepackage{makecell}
\usepackage{cuted}
\usepackage{caption}
\usepackage{tikz}
\usepackage[hidelinks]{hyperref}

\title{\LARGE \bf Tri-Manual Visuomotor Imitation Learning of Robot Policies
}

\author{%
\authorblockN{James Zhao$^{1,\dagger}$, Mingyuan Ba$^{1,\dagger}$, and Weiming Zhi$^{1,2,3,*}$}
\authorblockA{\footnotesize
$^{1}$School of Computer Science and $^{2}$Australian Center For Robotics, The University of Sydney, Australia\\
$^{3}$College of Connected Computing, Vanderbilt University, TN, USA\\
$^{\dagger}$Equal contribution. $^{*}$Corresponding author: \texttt{\scriptsize Weiming.Zhi@sydney.edu.au}.\\
Project website: \url{https://aus.bot/research/trimanpolicy/}}
}

\begin{document}

\maketitle
\thispagestyle{empty}
\pagestyle{empty}

\begin{strip}
\centering
\includegraphics[width=\linewidth]{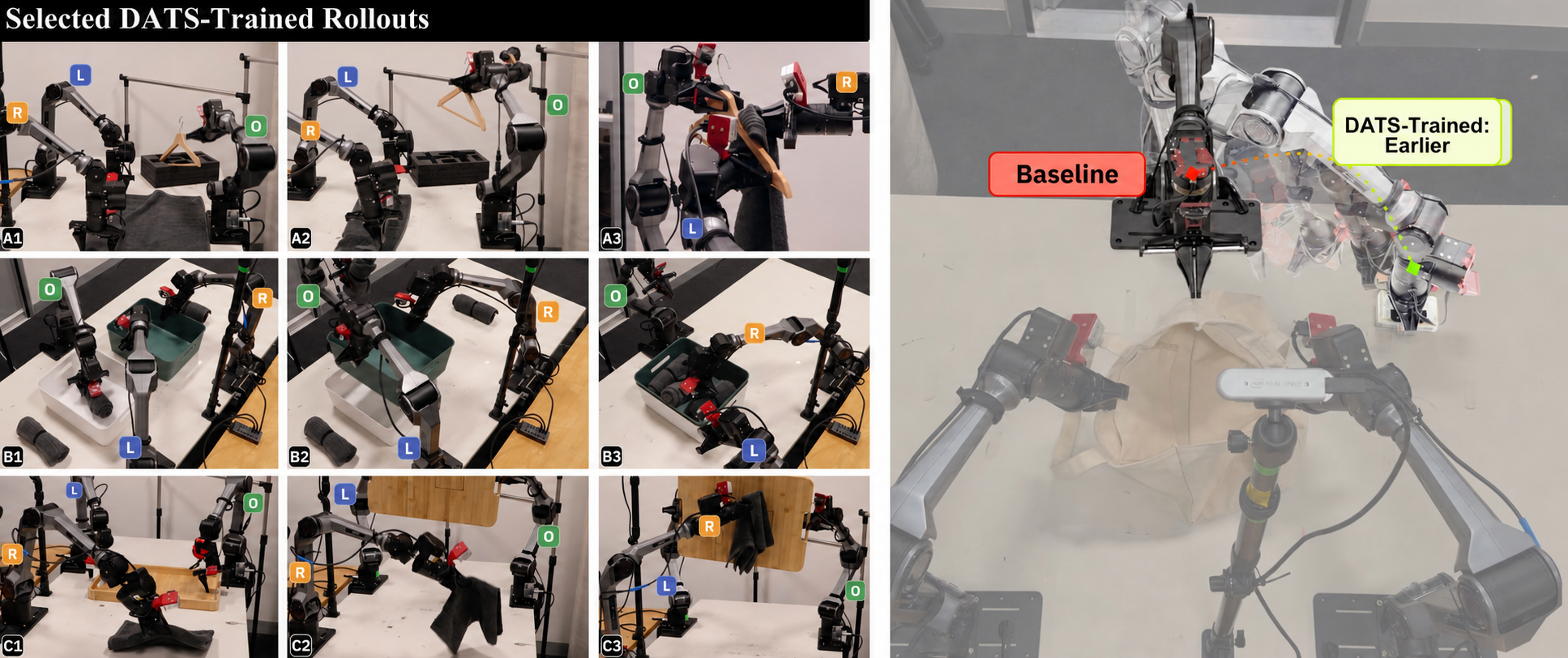}
\captionof{figure}{TriManPolicy learns synchronous three-arm visuomotor policies from pairwise demonstrations collected by one operator. Pairwise control can serialise independent actions; DATS reconstructs their global timing before policy training. Selected rollouts show the auxiliary arm maintaining task geometry while the other arms manipulate.}
\label{fig}
\end{strip}

\begin{abstract}
Bimanual teleoperation provides an effective way to collect robot demonstrations, but it assumes that the operator and robot have matching numbers of simultaneous control channels. This assumption breaks for tri-manual systems: the robot can coordinate three arms concurrently, whereas a single operator can continuously control only two. Pairwise mode switching may therefore record otherwise independent motions sequentially, causing behaviour cloning to reproduce delays imposed by the interface rather than required by the task. We present \emph{TriManPolicy}, a tri-manual imitation learning system that allows one operator to demonstrate behaviours for three arms. Its central component is Dependency-Aware Tri-Arm Scheduling (DATS). The key idea is to preserve the demonstrated arm motions while reconsidering when they occur. DATS retimes demonstrations offline by preserving fixed-duration local sensorimotor segments and repositioning them under task-dependency and arm-resource constraints. The resulting data train a single synchronous policy for all three arms, while deployment requires neither the dependency graph nor the scheduler. Across six challenging real-world tasks, DATS-trained policies have lower mean completion time among successful trials on every task, with comparable observed success counts. Offline analysis shows that DATS changes which cross-arm targets enter common action-chunk windows.
\end{abstract}

\section{Introduction}
Direct teleoperation has enabled recent advances in visuomotor imitation learning by turning human manipulation into robot observation-action trajectories. This correspondence is natural for bimanual systems, where an operator's two hands map to two robot end effectors~\cite{zhao2023learning,fu2024mobilealoha}. However, it also embeds an implicit assumption: the operator and robot expose the same number of simultaneous control channels. We study teleoperated imitation learning beyond this matched setting, where the robot's manipulation capabilities exceed those of the demonstration interface.

Many manipulation tasks can benefit from distributing complementary roles across more than two end effectors. While two arms perform the primary manipulation, a third can stabilise, support, or present an object without interrupting their motion. For example, it can hold a bag open during insertion or support a hanger while two arms arrange a towel. With only two manipulators, one arm may need to alternate between maintaining support and advancing the task, introducing additional regrasping, waiting, or sequential execution. A tri-manual embodiment avoids this trade-off by providing a third simultaneous manipulation channel. However, it also breaks the correspondence assumed by standard bimanual teleoperation: the robot can coordinate three arms concurrently, whereas the operator can continuously control only two.

Pairwise mode switching allows one operator to demonstrate behaviours for all three arms, but only two at a time. As Fig.~\ref{fig} shows, motions that could occur concurrently may therefore be recorded sequentially. Behaviour cloning can absorb this staging imposed by the interface and reproduce unnecessary waiting at deployment. The key observation is that the demonstrated motions may remain useful even when their recorded timing does not. Dependencies required by the task should therefore be preserved, while timing introduced only by the interface should not automatically become part of the learning target. Motivated by this observation, we introduce \emph{TriManPolicy}, a tri-manual imitation learning system that separates demonstrated motion from collection timing while allowing a single operator to provide the demonstrations. Its core component, Dependency-Aware Tri-Arm Scheduling (DATS), reconstructs the global timing of demonstrated motions under task-dependency and arm-resource constraints. These retimed demonstrations train a synchronous visuomotor policy that controls all three arms jointly.

This work makes three contributions:
\begin{itemize}
\item We extend teleoperated imitation learning to a setting in which the robot can act through more simultaneous manipulation channels than a single operator can demonstrate, and formulate this problem for three cooperating arms.
\item We introduce DATS, which casts demonstration retiming as a constrained scheduling problem. Independent arm motions may overlap, while required task orderings and constraints on arm usage are preserved.
\item We train synchronous policies that jointly control all three arms and evaluate the complete system across six challenging tasks in the real world using both policy rollouts and offline analyses of coordination.
\end{itemize}

\section{Related Work}

\noindent\textbf{Visuomotor imitation learning:} Learning from demonstration is a standard route to manipulation skills when rewards are difficult to specify~\cite{argall2009survey,ravichandar2020robot}. Structured LfD models can also adapt demonstrated motions to new environments while retaining stable execution~\cite{zhi2022diffeomorphic}. Action-chunked transformer policies, diffusion policies, and bimanual systems learn directly from recorded visuomotor trajectories~\cite{zhao2023learning,chi2023diffusion,fu2024mobilealoha,zhao2025alohaunleashed}. These methods generally retain the recorded global timeline as the supervision target. TriManPolicy studies what that timeline should contain when the collection interface is less parallel than the robot embodiment.

\noindent\textbf{Demonstration interfaces and multi-arm manipulation:} Demonstration interfaces shape both the data distribution and its timing. Recent work studies portable hardware, in-the-wild datasets, and generated demonstrations~\cite{chi2024umi,khazatsky2024droid,mandlekar2023mimicgen}. Other approaches derive motion supervision from human video, sketches, or cross-modal annotations~\cite{dong2026jointflow,zhi2024diagrammatic,zhi2025periodic,barron2026crossmodal}. Bimanual systems support stabilisation and handover~\cite{smith2012dual}. MART learns two- and three-arm coordination from simultaneous multi-user teleoperation~\cite{tung2020mart}, whereas TriManPolicy uses one operator with pairwise continuous control. Tri-manual and supernumerary-limb work provides further examples of an additional end-effector supporting manipulation~\cite{huang2020trimanipulation,huang2023threearms,tong2021srl}.

\noindent\textbf{Temporal structure and scheduling:} Task graphs and temporal decompositions support skill learning, task segmentation, and long-horizon planning~\cite{konidaris2012skill,ma2023adaptive,shiarlis2018taco,dreher2024temporal}, while constraint scheduling is widely used to plan multi-robot and multi-arm execution~\cite{behrens2019constraint,pan2021multiple,chen2022cooperative}. DATS applies this scheduling perspective to the construction of behaviour-cloning supervision from recorded sensorimotor segments.

\noindent\textbf{Closest contemporaneous approaches:} HATS assigns assistive arms to an online multimodal agent so that one operator need not mode-switch~\cite{lin2026hats}. Sequential Asymmetric Imitation (SAI) uses a staged single-teleoperator curriculum to train physically coupled robot policies against increasingly realistic partner behaviour~\cite{chen2026sai}. DexImit schedules subtasks extracted from monocular human videos and then synthesises bimanual robot trajectories~\cite{mu2026deximit}. Dreher \emph{et al.} learn temporal constraints for executable bimanual plans~\cite{dreher2026unified}, and Proleptic Temporal Ensemble changes deployment-time action aggregation to accelerate an imitation policy~\cite{park2024proleptic}. DATS keeps the executed robot segments and changes when they appear in training supervision.

\begin{figure}[t]
\centering
\includegraphics[width=\linewidth]{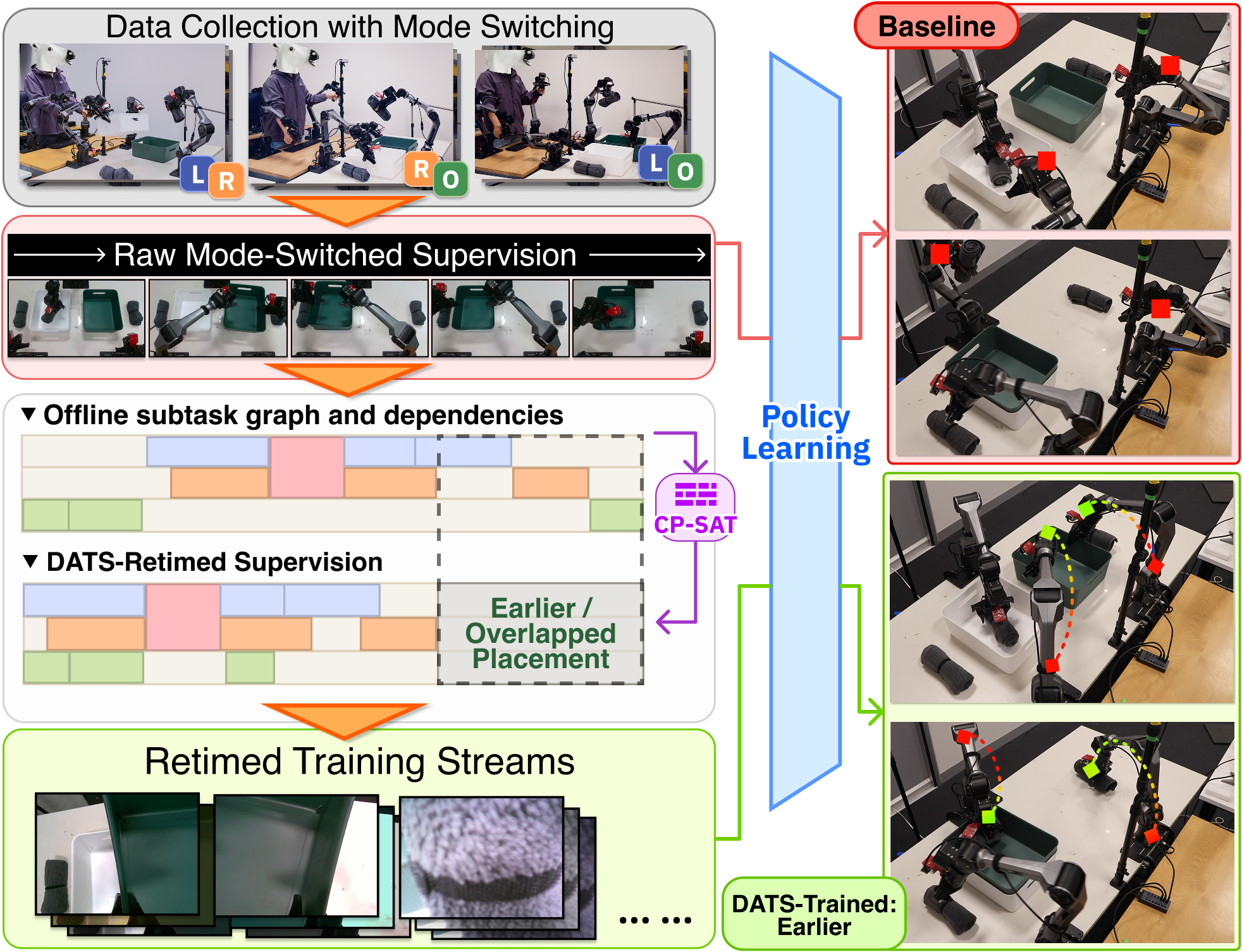}
\caption{TriManPolicy pipeline. An offline annotation pass constructs an episode-level subtask graph from each mode-switched demonstration; DATS uses the graph to optimise segment placement and produce retimed three-arm supervision. Raw and DATS share the policy-learning module.}
\label{fig:method_pipeline}
\end{figure}

\section{TriManPolicy System and Method}

\subsection{System Overview and Problem Formulation}

TriManPolicy combines mode-switched demonstration collection, offline DATS supervision construction, and a synchronous three-arm visuomotor policy. Figure~\ref{fig:method_pipeline} gives the full pipeline. Each recorded episode is segmented into local behaviours, and an annotation pass assigns required arms and predecessor relations. DATS then chooses new start times for these fixed-duration segments. The task graph separates order that the task requires from order that the scheduler may relax. Predecessor edges encode finish-to-start relations, including task prerequisites and selected shared-workspace orderings. Required-arm sets identify resources that cannot be used concurrently. A valid schedule satisfies both conditions while minimising episode makespan. Segment samples and their local sensorimotor timing stay fixed, while DATS re-optimises their placement on the supervision timeline.

The scheduled arm-centric streams form a new observation and action sequence. Retiming changes episode duration, held values, and which arm behaviours occur together inside action-chunk training windows. This offline data transformation rearranges recorded samples during dataset construction, while the learned policy acts directly from observations. A mode-switched demonstration can contain the right local motions yet present them on the wrong global clock: an arm waits because the operator is controlling another pair, not because the task requires a delay. DATS keeps the demonstrated motions and replaces that clock with one defined by task relations and available arm resources. Static support remains part of the supervision when it maintains a task condition; DATS schedules graph-permitted concurrency while preserving required holds.

\subsection{Mode-Switched Demonstration Collection}

\begin{figure}[t]
\centering
\includegraphics[width=\linewidth]{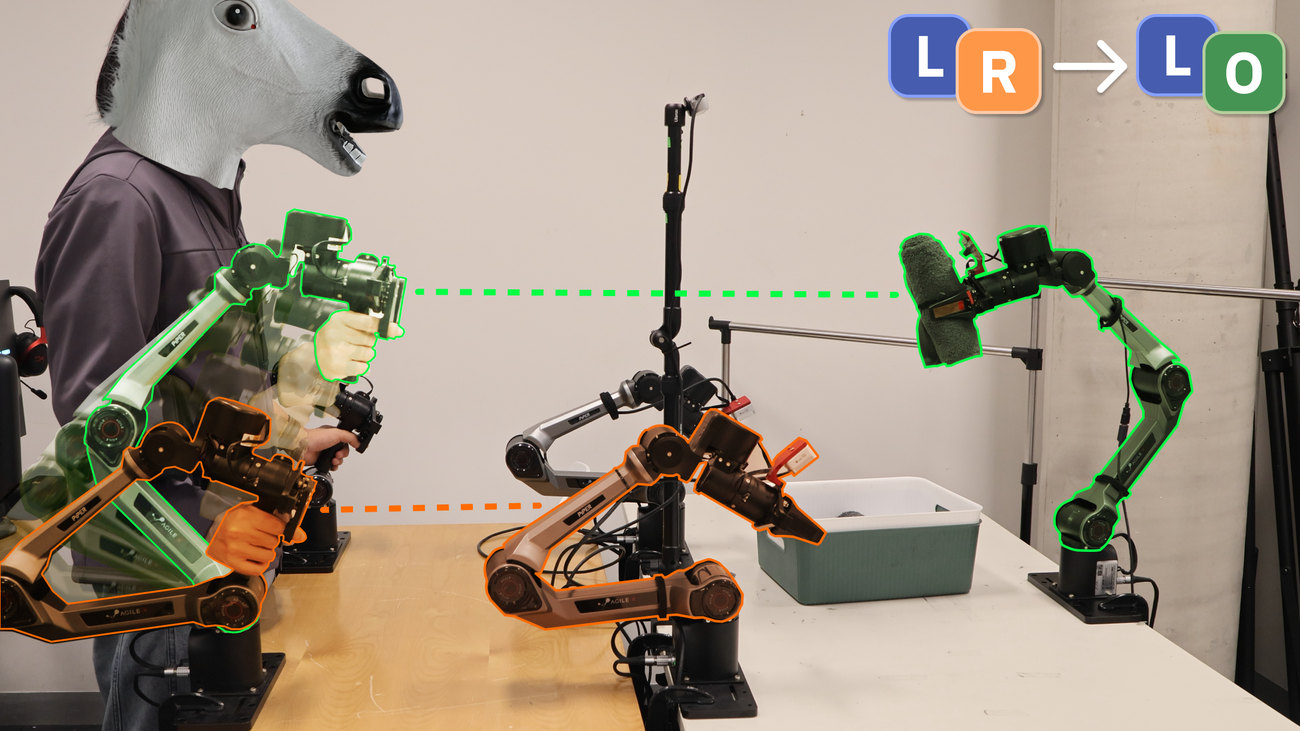}
\caption{Mode-switched teleoperation interface for three-arm demonstration collection.
The interface supports pairwise control assignments over Arms L, R, and O, allowing one operator to collect three-arm demonstrations while controlling only two arms at a time.}
\label{fig:teleop_mode_switch}
\end{figure}

We collect demonstrations using the mode-switched interface in Fig.~\ref{fig:teleop_mode_switch}.
The operator provides two continuous input streams, assigned at each timestep to one of the arm pairs $\{\mathrm{LR},\mathrm{LO},\mathrm{RO}\}$ over robot arms $\mathcal A=\{L,R,O\}$.
This enables one operator to collect three-arm demonstrations without simultaneous three-arm teleoperation. Changing the active pair is a discrete mode switch, and the unselected arm remains under hold control.
All robot states, gripper states, teleoperation modes, observations, and executed commands are logged under a unified control clock:
\begin{align}
\mathcal D_{\mathrm{raw}}
&=\{\tau_n^{\mathrm{raw}}\}_{n=1}^{N}, \notag\\
\tau_n^{\mathrm{raw}}
&=
\{(o_{n,t}^{\mathrm{raw}},a_{n,t}^{\mathrm{raw}},m_{n,t})\}_{t=1}^{T_n}, \notag\\
a_{n,t}^{\mathrm{raw}}
&=
[u_{n,t}^{L,\mathrm{raw}},u_{n,t}^{R,\mathrm{raw}},u_{n,t}^{O,\mathrm{raw}}] .
\end{align}
The observation $o_{n,t}$ contains arm-centric images and proprioception, while $a_{n,t}$ concatenates commands for all three arms. Commands are sampled causally from the latest executed command at or before the observation timestamp. The mode trace $m_{n,t}$ records the active teleoperation pair for data inspection; DATS operates on the annotated segments and episode graph.

\subsection{Subtask Annotation and Graph Construction}

For each raw episode, a VLM proposes segment boundaries, required arms, semantic labels, and predecessor relations from the video. Before scheduling, an author reviews these structural annotations and edits them where needed, focusing on object-state prerequisites and intervals that establish or maintain support. DATS treats the resulting graph $\mathcal G_n=(\mathcal V_n,\mathcal E_n)$ as the episode-level task specification. Automated checks reject malformed or cyclic graphs and invalid durations or arm identifiers; the scheduler then enforces the predecessor and resource constraints encoded by that specification.

Semantic labels describe the segments but do not enter the scheduling objective. Only interval boundaries, required-arm sets, and predecessor relations affect the schedule. Each predecessor relation has finish-to-start semantics and may require a support state to be established before a consumer segment begins. The interval records the active motion and arm occupancy.

The segments are episode-specific temporal units. A segment may use one arm or several, and its duration is taken from the recorded execution. The annotation specifies when the segment occurred, which arms it occupied, and which earlier segments must finish before it begins. This compact representation supplies the temporal and resource information needed to retime the demonstration while leaving the sensorimotor content in the data.

Each node $v_j\in\mathcal V_n$ corresponds to a temporally localised subtask and is represented as
$v_j=(\mathcal A_j,\bar s_j,\bar e_j,\mathcal P_j,y_j)$, where $\mathcal A_j$ is the required-arm set, $\bar s_j$ and $\bar e_j$ are raw start and end times, $\mathcal P_j$ is the predecessor set, and $y_j$ is an optional semantic label.
The directed edge set is induced by the predecessor sets:
\begin{equation}
\mathcal E_n
=
\{(v_p,v_j)\mid v_p\in\mathcal P_j,\; v_j\in\mathcal V_n\}.
\end{equation}

The graph $\mathcal G_n$ encodes all predecessor relations enforced by the implemented scheduler. For each arm $\alpha\in\mathcal A$, let $\mathcal V_n^\alpha$ denote the subtasks that involve arm $\alpha$:
\begin{equation}
\mathcal V_n^\alpha
=
\{v_j\in\mathcal V_n\mid \alpha\in\mathcal A_j\}.
\end{equation}
The scheduler prevents intervals in $\mathcal V_n^\alpha$ from overlapping, but it does not independently impose their raw order. Arm-local transitions and cross-arm object-state prerequisites are therefore encoded in the predecessor sets $\mathcal P_j$. Required-arm sets describe occupancy within an interval.

Separating precedence from resource occupancy lets the graph express two different reasons for serial execution. A predecessor edge preserves a task relation even when two segments use different arms. A shared required arm prevents overlap even when no explicit edge connects the segments. Conversely, disjoint-arm segments without a dependency remain free to move relative to one another.

The predecessor sets need not reproduce raw chronology. To formalise the motivating mismatch, we identify raw-ordered, non-overlapping segment pairs that have no dependency path and use disjoint arms:
\begin{align}
\mathcal C_n
=
\{(v_i,v_j)\mid\;&
v_i\prec_{\mathrm{raw}} v_j,\;
v_i\not\rightsquigarrow v_j,\;
v_j\not\rightsquigarrow v_i, \notag\\
&
\mathcal A_i\cap\mathcal A_j=\emptyset
\}.
\end{align}
Here, $v_i\prec_{\mathrm{raw}}v_j$ iff $\bar e_i\leq\bar s_j$, and $v_i\rightsquigarrow v_j$ denotes that there exists a directed dependency path from $v_i$ to $v_j$ in $\mathcal G_n$.
The set $\mathcal C_n$ identifies raw cross-arm orderings that the graph permits the scheduler to relax and is later used to audit target-window composition. It adds no scheduling constraint; the task graph and arm resources determine feasibility.

\subsection{Dependency-Aware Tri-Arm Scheduling}

\begin{figure}[t]
    \centering
    \begin{tikzpicture}
        \node[anchor=south west,inner sep=0] (schedule) at (0,0)
        {\includegraphics[width=\linewidth]{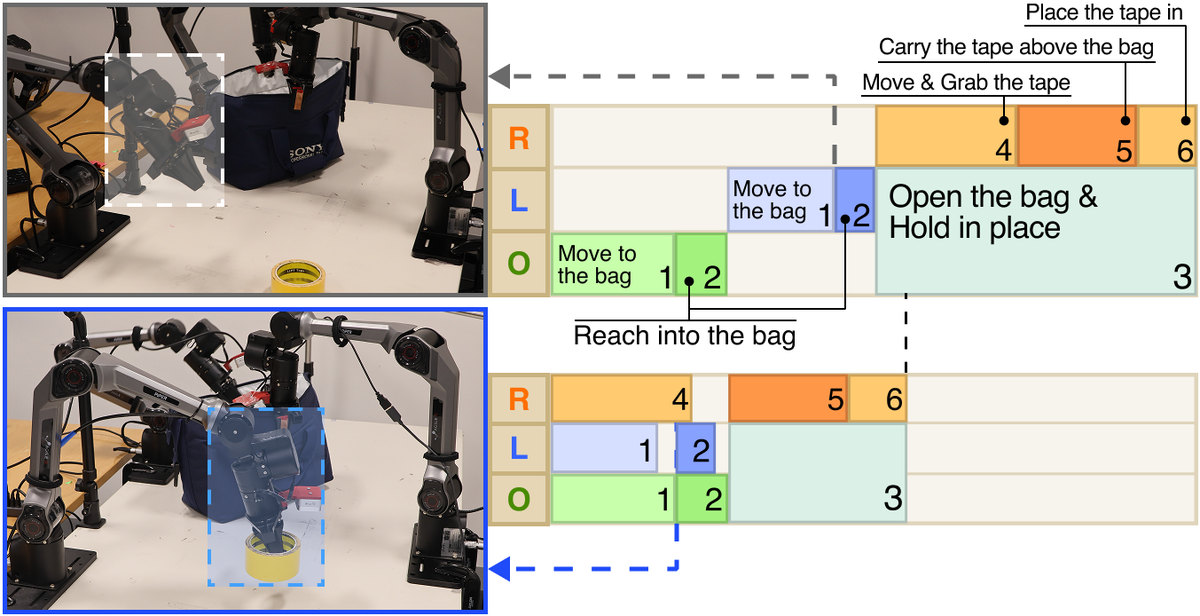}};
        \begin{scope}[x={(schedule.south east)},y={(schedule.north west)}]
            \fill[purple,opacity=0.12] (0.50,0.52) rectangle (0.59,0.83);
            \draw[purple!85!black,densely dashed,line width=0.8pt]
                (0.50,0.52) rectangle (0.59,0.83);
            \node[anchor=south,font=\tiny\bfseries,fill=white,inner sep=0.5pt,
                  text=purple!85!black] at (0.545,0.835) {Raw target window};
            \fill[purple,opacity=0.12] (0.50,0.15) rectangle (0.59,0.39);
            \draw[purple!85!black,densely dashed,line width=0.8pt]
                (0.50,0.15) rectangle (0.59,0.39);
            \node[anchor=north,font=\tiny\bfseries,fill=white,inner sep=0.5pt,
                  text=purple!85!black] at (0.545,0.145) {DATS target window};
        \end{scope}
    \end{tikzpicture}
    \caption{
    DATS temporal reparameterisation on \textsc{BagTape}.
    Equal-width purple bands schematically mark horizon-$H$ action-target windows.
    The Raw window contains bag preparation on O with held targets on L and R; the DATS window presents tape handling on R together with bag preparation on L and O.
    The aperture-insertion dependency remains enforced.
    }
    \label{fig:bagtape_scheduling_example}
\end{figure}

We formulate temporal reparameterisation as \emph{Dependency-Aware Tri-Arm Scheduling} (DATS), a fixed-duration interval scheduling problem over three robot arms. Let $W_t^H=\{t,t+1,\ldots,t+H-1\}$ denote the sample indices covered by a horizon-$H$ target beginning at $t$. Figure~\ref{fig:bagtape_scheduling_example} illustrates the scheduling idea on \textsc{BagTape} and marks the corresponding change in such a target window. Arm R can handle the tape while Arms L and O prepare the bag because these segments use disjoint arms and have no precedence path. Tape insertion still follows aperture formation. Relaxing a cross-arm ordering on the critical path shortens the supervision timeline; elsewhere it changes when task-relevant segments appear even if total duration stays similar.

Time is discretised into ticks of length $\Delta t=0.01\,\mathrm{s}$. For
each subtask, we convert its raw interval into tick indices, retain the
resulting duration, and represent its scheduled placement as
\begin{equation}
\begin{aligned}
(\hat s_j,\hat e_j)
&=
\left(
\left\lfloor\frac{\bar s_j}{\Delta t}\right\rfloor,
\left\lceil\frac{\bar e_j}{\Delta t}\right\rceil
\right),\\
d_j&=\hat e_j-\hat s_j,\\
e_j&=s_j+d_j,\qquad \mathcal T_j=[s_j,e_j).
\end{aligned}
\end{equation}
Here, $s_j$ and $e_j$ are the scheduled start and end ticks.

The schedule satisfies two constraint types. First, the specified predecessor relations are enforced:
\begin{equation}
e_p\leq s_j,
\qquad
\forall v_j\in\mathcal V_n,
\forall v_p\in\mathcal P_j .
\end{equation}
Second, each robot arm can execute at most one subtask at a time. Let
\begin{equation}
\mathcal Q_\alpha=\{\mathcal T_j\mid \alpha\in\mathcal A_j\},
\qquad
\alpha\in\mathcal A .
\end{equation}
We impose $\mathrm{NoOverlap}(\mathcal Q_\alpha)$ for each arm $\alpha\in\mathcal A$; a multi-arm subtask occupies all required arms for its full scheduled interval. A terminal support pose may persist through the inactive-arm hold used in construction, provided the predecessor relations prevent incompatible later use of that arm.
For example, in \textsc{BagTape}, bag opening has $\mathcal A_6=\{L,O\}$, tape insertion has $\mathcal A_9=\{R\}$, and $(v_6,v_9)\in\mathcal E_n$. L and O have no later segment, so their final aperture pose is held throughout insertion; the edge establishes the aperture, while construction supplies its maintenance.

We optimise the schedule by minimising the episode makespan:
\begin{equation}
\begin{aligned}
&\{s_j^\ast,e_j^\ast\}_{v_j\in\mathcal V_n}, C_{\max}^\ast
\in
\arg\min_{\{s_j,e_j\},C_{\max}}
\quad C_{\max} \\
\mathrm{s.t.}\quad
& C_{\max}\geq e_j, \quad v_j\in\mathcal V_n, \\
& e_j=s_j+d_j, \quad v_j\in\mathcal V_n, \\
& e_p\leq s_j, \quad v_j\in\mathcal V_n,\ v_p\in\mathcal P_j, \\
& \mathrm{NoOverlap}(\mathcal Q_\alpha), \quad \alpha\in\mathcal A, \\
& s_j,e_j,C_{\max}\in\mathbb Z_{\geq 0}.
\end{aligned}
\end{equation}

We solve this problem using CP-SAT in Google OR-Tools 9.12~\cite{cpsatlp} and denote the returned schedule by $\mathcal S_n^\ast=\{(s_j^\ast,e_j^\ast)\}_{v_j\in\mathcal V_n}$. It satisfies the specified predecessor order and arm-resource exclusivity and is stored as the episode's retiming map. Other placements may attain the same optimal makespan away from the critical path; construction and audit use this stored map throughout.

Makespan minimisation creates overlap where the graph and resource constraints permit it, while long support intervals can remain on the critical path. Because these quantities need not move together, we examine duration, segment placement, and target-window composition separately.

\subsection{Constructing Retimed Demonstrations}

For each raw message timestamp $q\in[\bar s_j,\bar e_j)$, DATS applies the forward time map
\begin{equation}
\phi_{n,j}(q)=\Delta t\,s_j^\ast+(q-\bar s_j).
\end{equation}
All messages for a participating arm are shifted by the same offset, and multi-arm segments share it. The retimed streams are sampled on the dataset clock: images and proprioception use the nearest timestamped sample, while commands use the latest sample at or before each output time. Initial values are used before an arm's first interval, and the most recent values are held between intervals and after the last one. Tick rounding may extend an interval by up to two ticks, in which case its final value is held to the scheduled end. Construction uses timestamp shifts, resampling, and held values rather than synthesising transitions at segment boundaries.

The resulting observations can combine arm-centric streams recorded at different raw times. Within each participating arm and multi-arm segment, the demonstrated sensorimotor timing remains intact. This construction assumes that graph-compatible arm-centric segments remain locally meaningful when composed. At deployment, the policy receives live, physically synchronous observations. The closed-loop robot trials test whether training on these composites yields executable three-arm behaviour.

For a multi-arm segment, every participating stream receives the same temporal offset, preserving the coordination recorded inside that segment. Independently annotated segments can move relative to one another, which changes the composite observation and action target seen at a given training time.

Although introduced for three arms, the optimisation is defined over a finite resource set. Here, resources are manipulators; the same exclusivity constraint can also represent a tool or workspace region. A segment may occupy several resources while retaining its recorded within-segment coordination. Retiming is appropriate when the interface serialises behaviours that deployment can execute concurrently, the segments remain locally meaningful after they move, and the task graph captures the state prerequisites between them. The current system instantiates this formulation with three arms, CP-SAT scheduling, and an action-chunked policy.

\begin{figure*}[t]
\centering
\includegraphics[width=\textwidth]{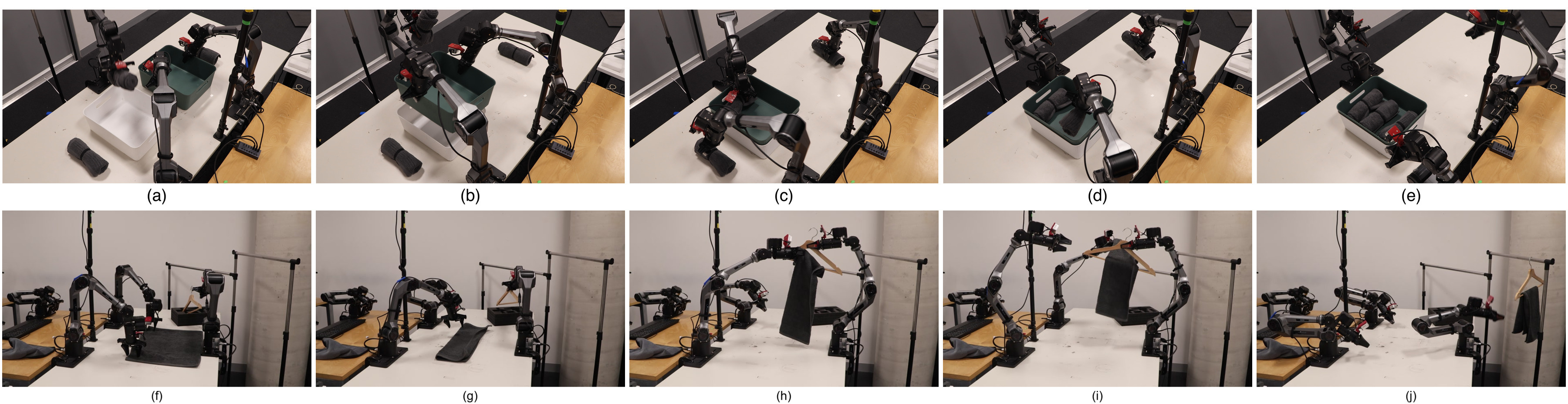}
\caption{Selected successful DATS rollouts for \textsc{BinTowel} (top) and \textsc{TowelHang} (bottom). In \textsc{BinTowel}, O moves the green container as L and R handle the towels. In \textsc{TowelHang}, O retrieves the hanger as L and R fold and lift the towel.}
\label{fig:selected_rollout_sequences}
\end{figure*}

\subsection{Policy Learning and Deployment}

For both datasets, we use an action-chunked transformer policy. Given observation $o_t$, the policy predicts a horizon-$H$ sequence of synchronised three-arm actions, $\hat A_{t:t+H-1}=\pi_\vartheta(o_t)$. The policy runs continuously across segment boundaries.

\begin{samepage}
Let $\Omega_H(\mathcal D)$ be the deterministic index set of all valid observation--action windows constructed from demonstration set $\mathcal D$, so that each $(n,t)\in\Omega_H(\mathcal D)$ identifies $(o_{n,t},A_{n,t}^H)$. Training minibatches sample these indexed windows. Raw and DATS use the same terminal-padding and loss-masking rule; padded indices receive no segment membership in the offline audit. Both training conditions minimise
\begin{equation}
\mathcal L(\vartheta;\mathcal D)
=
\frac{1}{|\Omega_H(\mathcal D)|}
\sum_{(n,t)\in\Omega_H(\mathcal D)}
\ell\!\left(\pi_\vartheta(o_{n,t}),A_{n,t}^H\right),
\end{equation}
\end{samepage}
where $A_{n,t}^H=[a_{n,t},\ldots,a_{n,t+H-1}]$. Although both conditions optimise the same objective, they produce different horizon-$H$ windows. Raw mode switching can pair progress by one arm with interface-induced holds by the others. DATS preserves each segment's local sensorimotor sequence but places compatible cross-arm segment targets within the same window, changing the joint supervision presented to the synchronous policy. Makespan alone does not capture this change: two timelines of similar duration can contain different joint three-arm targets at the same horizon. The co-window coverage diagnostic in Sec.~IV-C measures this change in the supervision directly. At deployment, the policy maps live observations directly to synchronised actions.

\section{Experiments}
\label{sec:experiments}

We evaluate TriManPolicy on six real-world three-arm tasks using 237 processed demonstrations. For each task, we ran 25 real-robot trials per training condition, giving 300 trials in total. We first report the robot outcomes, then analyse how DATS changes episode duration and the composition of the policy's target windows. Figure~\ref{fig:selected_rollout_sequences} previews two complementary forms of learned coordination: in \textsc{BinTowel}, O stages the green container while L and R continue towel handling; in \textsc{TowelHang}, O brings the movable hanger into the workspace while L and R fold and lift the towel.

\begin{table}[t]
\centering
\scriptsize
\setlength{\tabcolsep}{1.4pt}
\renewcommand{\arraystretch}{1.06}
\begin{tabular}{@{}p{0.18\columnwidth} p{0.36\columnwidth} p{0.39\columnwidth}@{}}
\toprule
Task & Coupled stage & Completion criterion \\
\midrule
\textsc{TowelHang} & O holds the hanger; L/R manipulate the towel & Folded towel hung by O on the rear hanger \\
\textsc{BinTowel} & O stages the container; L/R transfer towels & Green container on white; all towels inside green \\
\textsc{ToteCards} & L/R hold the tote open; O inserts the box & Card box inside the maintained-open tote \\
\textsc{BagTape} & L/O hold the bag open; R inserts tape & Tape inside the bag \\
\textsc{LidEraser} & O positions the lid; L/R place erasers & Both erasers inside; lid closed on the box \\
\textsc{TrayWipe} & L/O lift and stabilise the tray; R wipes & Prescribed underside wipe completed \\
\bottomrule
\end{tabular}
\caption{Complementary three-arm stages and task-specific completion criteria.}
\label{tab:task_roles}
\end{table}

\subsection{Evaluation Design and Task Suite}

The experiments use three identical AgileX Piper-X arms with grippers. Each policy receives arm-centric RGB and proprioceptive observations and sends synchronous joint and gripper commands at 50~Hz. All six tasks require coupled three-arm behaviour: one arm establishes or maintains a condition needed by another. Table~\ref{tab:task_roles} summarises these stages and the completion criteria.

Each DATS episode is a retimed version of its Raw counterpart, so both conditions use the same 237 processed demonstrations. Three invalid \textsc{TowelHang} recordings were excluded, leaving 47 episodes for that task. The learner is also held fixed: an action-chunked transformer with a frozen SigLIP~2 Base-Patch16-256 encoder~\cite{tschannen2025siglip2}, a 100-step ($2.0\,\mathrm{s}$) horizon, and temporal aggregation. We train one policy per task and condition for 100k steps using batch size 16 and AdamW with a fixed $10^{-5}$ learning rate and seed 1000. The episode graphs contain 6 to 12 segments and 6 to 12 direct predecessor relations; CP-SAT solved all 237 to optimality under the 30~s limit.

Trials were interleaved within each task and used matched initial states with several-centimetre object-pose variation. In \textsc{BinTowel}, one towel began inside the white container and was removed before final placement. We measured completion from policy release after pre-alignment to the first visual satisfaction of the criterion in Table~\ref{tab:task_roles}. Safety stops or states unrecoverable for more than 10~s counted as failures, which entered the success counts but not mean completion time. Aggregate success pools all trials; the macro-mean time averages the six task means.

\subsection{System-Level Real-Robot Outcomes}

Table~\ref{tab:baseline_dats_results} reports the physical evaluation. DATS-trained policies have lower mean completion time among successful trials on every task, with reductions between $31.3\%$ and $49.9\%$. The unweighted macro-mean falls from $83.1\,\mathrm{s}$ to $47.8\,\mathrm{s}$, a $42.5\%$ reduction. DATS records 129 successes in 150 trials, compared with 126 for Baseline, with an equal or higher task-level count in all six tasks. DATS therefore combines lower successful-trial completion times with comparable observed success counts across the task suite.

\begin{table}[t]
\centering
\footnotesize
\setlength{\tabcolsep}{1.5pt}
\renewcommand{\arraystretch}{1.12}
\begin{tabular}{@{}l c cc cc c@{}}
\toprule
&
&
\multicolumn{2}{c}{Baseline}
&
\multicolumn{2}{c}{DATS}
&
\\
\cmidrule(lr){3-4}
\cmidrule(lr){5-6}
Task & $N$
& Succ. & Time (s)
& Succ. & Time (s)
& \makecell{Time red.\\(\%)} \\
\midrule
\textsc{TowelHang} & 47 & 17/25 & $150.4\!\pm\!4.7$ & 19/25 & $83.6\!\pm\!0.8$ & 44.4 \\
\textsc{BinTowel}  & 30 & 20/25 & $83.6\!\pm\!5.5$  & 21/25 & $57.5\!\pm\!5.8$ & 31.3 \\
\textsc{ToteCards} & 50 & 20/25 & $94.1\!\pm\!11.7$ & 20/25 & $50.2\!\pm\!5.0$ & 46.7 \\
\textsc{BagTape}   & 30 & 25/25 & $55.5\!\pm\!0.6$  & 25/25 & $31.4\!\pm\!1.1$ & 43.5 \\
\textsc{LidEraser} & 30 & 25/25 & $63.1\!\pm\!0.6$  & 25/25 & $31.6\!\pm\!0.8$ & 49.9 \\
\textsc{TrayWipe}  & 50 & 19/25 & $52.0\!\pm\!1.0$  & 19/25 & $32.4\!\pm\!1.5$ & 37.8 \\
\midrule
All / macro-mean & 237 & 126/150 & 83.1 & 129/150 & 47.8 & 42.5 \\
\bottomrule
\end{tabular}
\caption{Real-robot comparison over 25 trials per task and condition. $N$ denotes processed training demonstrations. Time is mean$\pm$sample SD over successful trials; failures count in Succ. but are excluded from time. The final row gives aggregate success and unweighted macro-mean time.}
\label{tab:baseline_dats_results}
\end{table}

The task-level consistency motivates a closer look at what changes in the supervision. We next separate changes in episode duration from changes in target-window composition.

\subsection{Supervision Duration and Target Composition}

We analyse duration and target-window composition across the Raw, Gap, and DATS timelines. Gap removes every maximal interval during which no annotated segment is active and applies the resulting monotone time map across the episode. This preserves the recorded ordering and overlap of annotated segments but does not enforce the task graph. Raw-to-Gap therefore isolates global inactivity, whereas Gap-to-DATS compares this order-preserving reference with the dependency-aware placement of the same segment samples. Gap is an offline diagnostic rather than a trained policy condition. Figure~\ref{fig:bagtape_scheduling_example} visualises the resulting window-level difference for one episode. Table~\ref{tab:timeline_diagnostics} reports episode duration, and Table~\ref{tab:cowindow_diagnostics} reports co-window coverage across the dataset. Together, these diagnostics distinguish a shorter episode from a change in which cross-arm targets enter the same training window.

\begin{table}[t]
\centering
\footnotesize
\renewcommand{\arraystretch}{1.10}
\setlength{\tabcolsep}{3.2pt}
\begin{tabular}{@{}l r c c c c@{}}
\toprule
Task & $N$ & Raw & Gap & DATS & Gap$-$DATS \\
\midrule
\textsc{TowelHang} & 47 & $94.8\!\pm\!7.0$ & $89.5\!\pm\!9.5$ & $69.7\!\pm\!11.0$ & $19.8\!\pm\!7.0$ \\
\textsc{BinTowel}  & 30 & $94.7\!\pm\!8.9$ & $79.4\!\pm\!9.1$ & $55.5\!\pm\!7.7$  & $23.9\!\pm\!5.8$ \\
\textsc{ToteCards} & 50 & $65.3\!\pm\!11.7$ & $65.3\!\pm\!11.7$ & $66.1\!\pm\!11.6$ & $-0.8\!\pm\!0.7$ \\
\textsc{BagTape}   & 30 & $49.8\!\pm\!5.9$ & $44.1\!\pm\!6.6$ & $38.2\!\pm\!6.8$  & $6.0\!\pm\!1.7$ \\
\textsc{LidEraser} & 30 & $47.5\!\pm\!5.1$ & $41.7\!\pm\!4.2$ & $29.0\!\pm\!2.7$  & $12.7\!\pm\!2.5$ \\
\textsc{TrayWipe}  & 50 & $48.0\!\pm\!2.9$ & $48.0\!\pm\!3.0$ & $47.9\!\pm\!3.0$  & $0.1\!\pm\!0.2$ \\
\bottomrule
\end{tabular}
\caption{Offline supervision duration over 237 processed episodes (mean$\pm$sample SD, s). Gap$-$DATS is paired by episode; positive values indicate shorter DATS timelines.}
\label{tab:timeline_diagnostics}
\end{table}

\begin{table}[t]
\centering
\footnotesize
\setlength{\tabcolsep}{3.4pt}
\renewcommand{\arraystretch}{1.10}
\begin{tabular*}{\columnwidth}{@{\extracolsep{\fill}}l r c c c@{}}
\toprule
Task & $N$ & Raw & Gap & DATS \\
\midrule
\textsc{TowelHang} & 47 & $9.0\!\pm\!14.8$ & $32.3\!\pm\!14.1$ & $98.8\!\pm\!6.0$ \\
\textsc{BinTowel}  & 30 & $2.9\!\pm\!3.0$  & $9.1\!\pm\!4.1$   & $50.5\!\pm\!8.7$ \\
\textsc{ToteCards} & 50 & $0.0\!\pm\!0.0$  & $0.0\!\pm\!0.0$   & $94.0\!\pm\!16.4$ \\
\textsc{BagTape}   & 30 & $0.3\!\pm\!1.7$  & $8.1\!\pm\!2.7$   & $24.4\!\pm\!2.7$ \\
\textsc{LidEraser} & 30 & $3.3\!\pm\!2.8$  & $9.8\!\pm\!2.3$   & $63.8\!\pm\!9.2$ \\
\textsc{TrayWipe}  & 50 & $9.5\!\pm\!1.8$  & $9.5\!\pm\!1.8$   & $71.2\!\pm\!10.6$ \\
\midrule
All / macro-mean & 237 & 4.2 & 11.5 & 67.1 \\
\bottomrule
\end{tabular*}
\caption{Eligible-pair co-window coverage $\rho_n^X$ over 237 processed episodes (episode mean$\pm$sample SD, \%). The final row is the unweighted task macro-mean.}
\label{tab:cowindow_diagnostics}
\end{table}

The duration audit shows additional compression beyond Gap for \textsc{TowelHang}, \textsc{BinTowel}, \textsc{BagTape}, and \textsc{LidEraser}. The other two tasks separate placement from episode length. \textsc{ToteCards} has a DATS timeline $0.8\!\pm\!0.7\,\mathrm{s}$ longer than Gap because DATS enforces a finish-to-start relation in the task graph that the raw overlap violates, yet card transfer begins $29.3\!\pm\!5.4\,\mathrm{s}$ earlier. In \textsc{TrayWipe}, the paired duration difference is $0.1\!\pm\!0.2\,\mathrm{s}$, while towel preparation and wiping begin $22.5\!\pm\!2.3\,\mathrm{s}$ and $15.1\!\pm\!1.4\,\mathrm{s}$ earlier. Tape preparation in \textsc{BagTape} likewise advances by $25.5\!\pm\!4.9\,\mathrm{s}$.

To connect these schedule changes to the learner, we audit the candidate pairs $\mathcal C_n$ defined in Sec.~III-C. For timeline $X\in\{\mathrm{Raw},\mathrm{Gap},\mathrm{DATS}\}$, let $\mathcal C_n^{H,X}\subseteq\mathcal C_n$ contain the eligible pairs for which at least one valid horizon-$H$ target window contains samples from both segments. The episode-level co-window coverage is
\begin{equation}
\rho_n^X
=
\frac{|\mathcal C_n^{H,X}|}{|\mathcal C_n|}.
\end{equation}

Each eligible pair contributes at most once.

We compute the fraction within each episode so that finer graph segmentation does not dominate the aggregate. All 237 episodes contain eligible pairs, giving 2,193 pairs in total. Thus, $\rho_n^X$ measures the breadth of eligible pairs presented within common target windows, not their frequency or concurrency during deployment.

\begin{figure}[t]
\centering
\includegraphics[width=\linewidth,trim=14bp 7bp 2bp 2bp,clip]{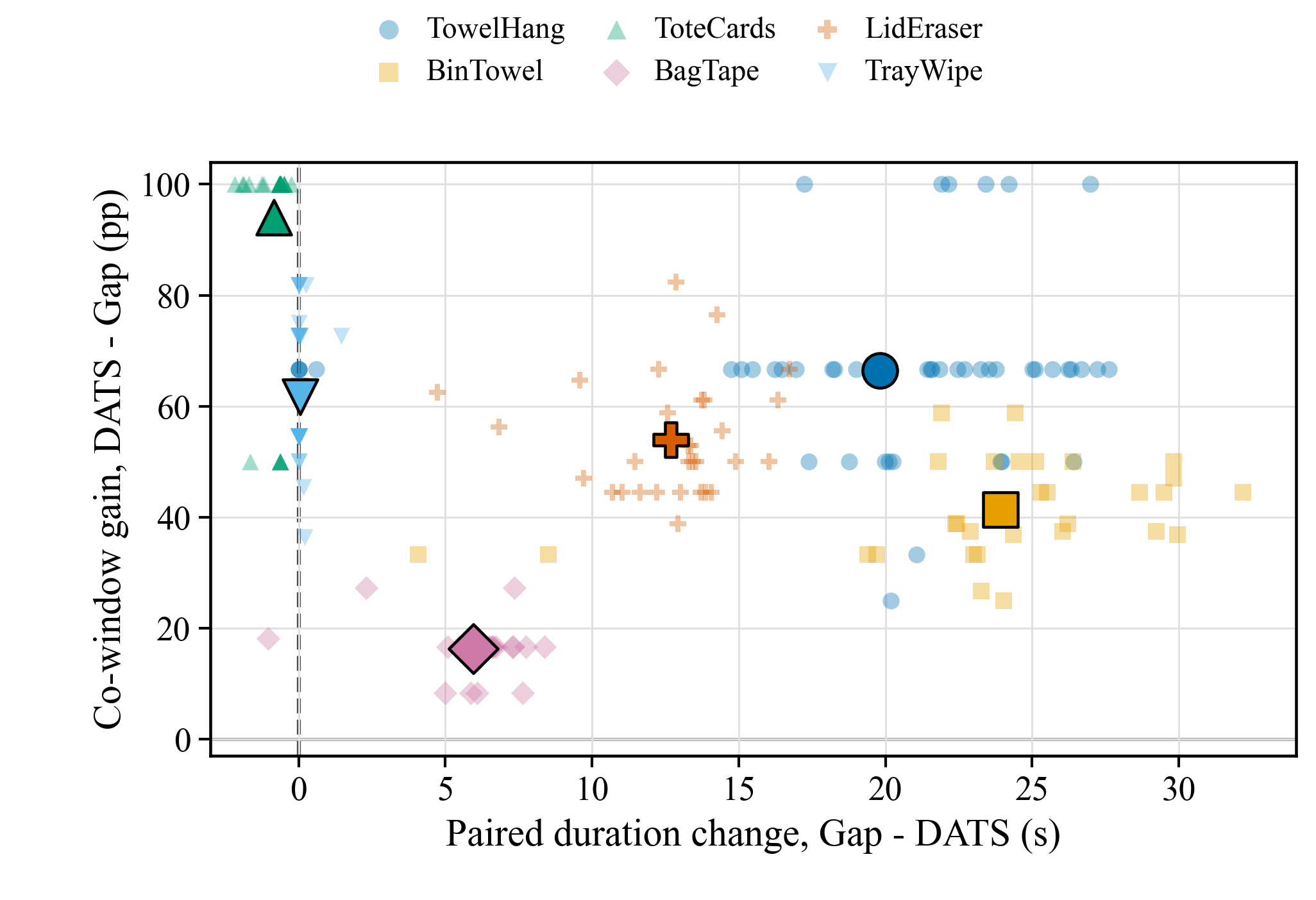}
\caption{Episode-level relationship between paired duration change and co-window gain over 237 processed demonstrations. Light markers show individual episodes; outlined markers show task means. Positive values correspond to shorter DATS timelines on the horizontal axis and greater DATS coverage on the vertical axis, measured in percentage points (pp).}
\label{fig:duration_cowindow_episode_audit}
\end{figure}

DATS increases co-window coverage over Raw and Gap on every task. The task macro-mean rises from $4.2\%$ in Raw and $11.5\%$ in Gap to $67.1\%$ under DATS. The paired comparison in Fig.~\ref{fig:duration_cowindow_episode_audit} shows higher co-window coverage under DATS than under Gap in all 237 processed episodes. The distinction between duration and coverage is clearest in \textsc{ToteCards} and \textsc{TrayWipe}. Each \textsc{ToteCards} episode contains two eligible pairs. Under Raw and Gap, the segments in each pair remain at least $8.6\,\mathrm{s}$ apart, giving zero coverage. DATS places at least one pair in a common $2.0\,\mathrm{s}$ target window in every episode, producing $94.0\!\pm\!16.4\%$ coverage even though its timeline is slightly longer. \textsc{TrayWipe} rises from $9.5\!\pm\!1.8\%$ under Raw and Gap to $71.2\!\pm\!10.6\%$ under DATS with nearly unchanged duration. In these two tasks, DATS changes target-window composition with little or no duration reduction; in the other four, broader coverage accompanies additional compression beyond Gap. The next section examines how coordination appears in policies trained on the transformed data.

\subsection{Coordination in Selected Rollouts}

\begin{figure}[t]
\centering
\includegraphics[width=\linewidth]{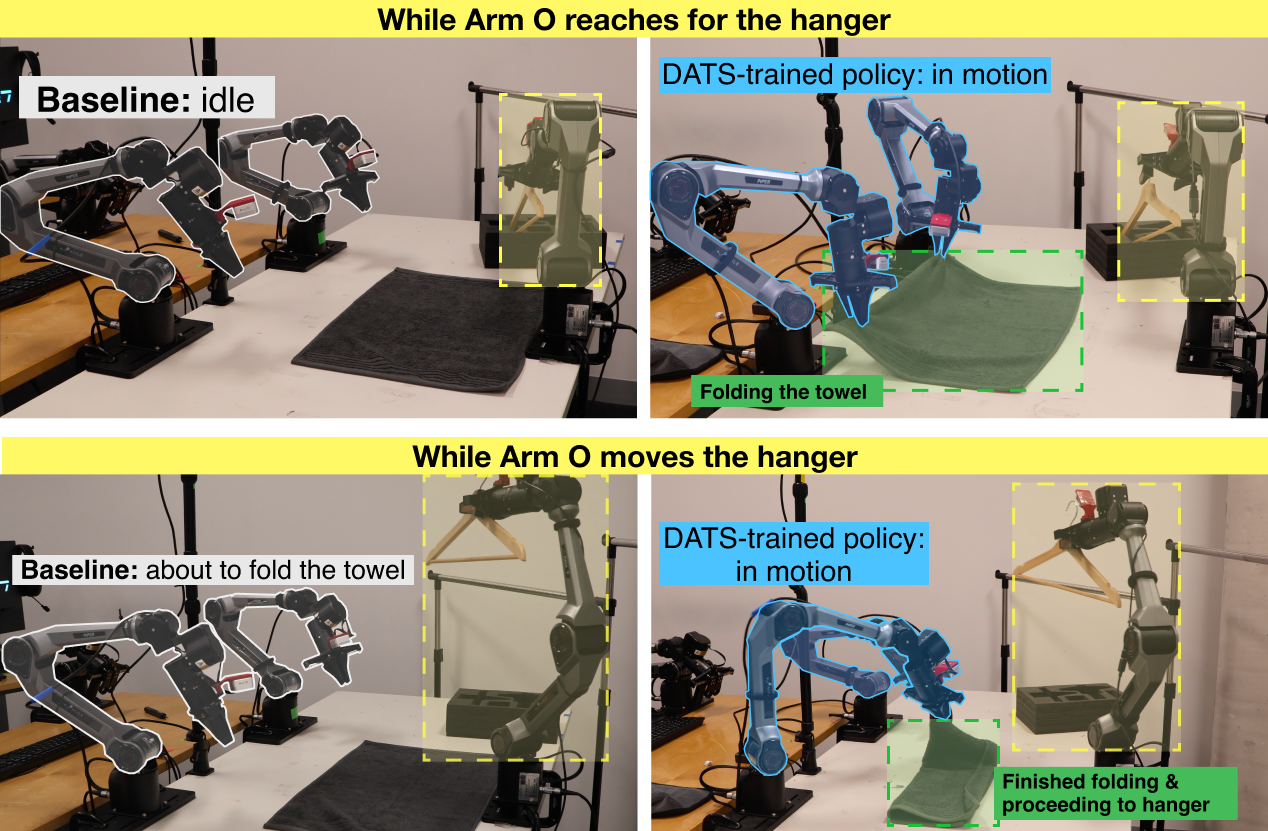}
\caption{Selected \textsc{TowelHang} comparison. While L and R fold and lift the towel, O remains idle under Baseline but retrieves the hanger under DATS.}
\label{fig:towelhang_three_arm_rollout}
\end{figure}

\begin{figure}[t]
\centering
\includegraphics[width=\linewidth]{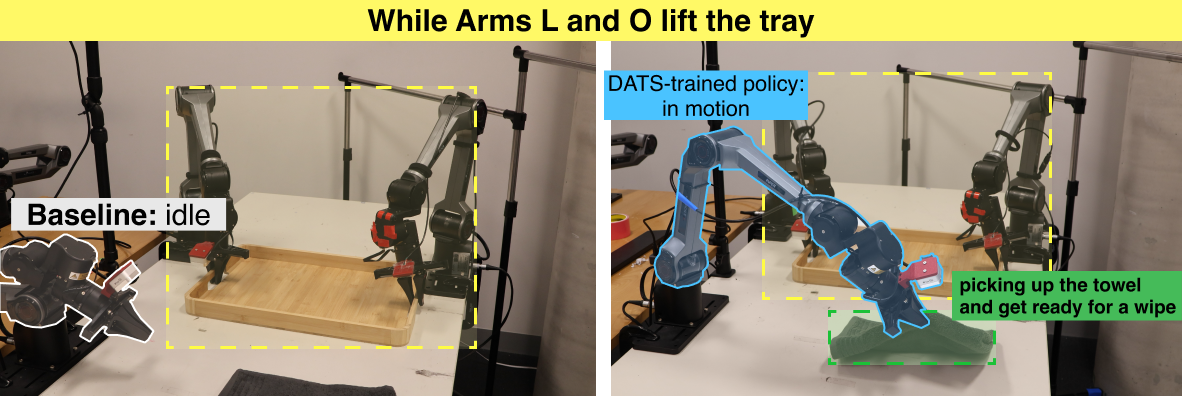}
\caption{Selected \textsc{TrayWipe} comparison. While L and O stabilise the tray, R remains idle under Baseline but begins towel pickup under DATS.}
\label{fig:traywipe_three_arm_rollout}
\end{figure}

\begin{figure}[t]
\centering
\includegraphics[width=\linewidth]{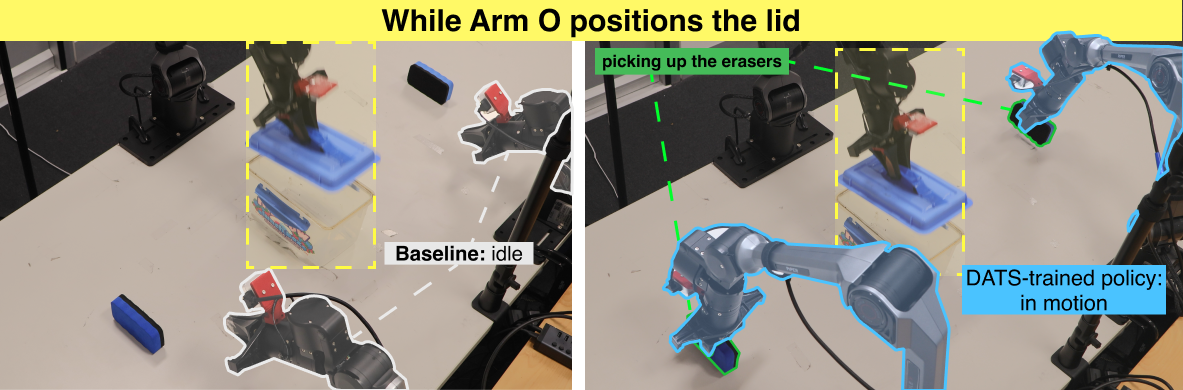}
\caption{Selected \textsc{LidEraser} comparison. While O positions the lid, L and R wait under Baseline but acquire the two erasers under DATS.}
\label{fig:lideraser_three_arm_rollout}
\end{figure}

Table~\ref{tab:baseline_dats_results} reports outcomes across all tasks and trials. Figure~\ref{fig:selected_rollout_sequences} provides end-to-end context for two successful DATS rollouts, whereas Figs.~\ref{fig:towelhang_three_arm_rollout}, \ref{fig:traywipe_three_arm_rollout}, and \ref{fig:lideraser_three_arm_rollout} isolate the stages at which DATS-trained policies advance while the corresponding Baseline policies wait. Hanger retrieval overlaps towel preparation, towel pickup begins during sustained tray support, and both erasers are acquired during lid positioning. Across these examples, required support persists while an independent stream begins earlier.

\subsection{Discussion}

Across the six separately trained task pairs, DATS is faster on successful trials and has an equal or higher observed success count. The 25-trial distributions characterise the evaluated policies across fixture placement, aperture formation, stabilisation, and object positioning. Because each condition uses one training run, they do not estimate variation across training seeds.

The offline audit separates two regimes. In \textsc{TowelHang}, \textsc{BinTowel}, \textsc{BagTape}, and \textsc{LidEraser}, broader co-window coverage accompanies shorter DATS timelines relative to Gap. In \textsc{ToteCards} and \textsc{TrayWipe}, compatible progress targets enter common action windows even though total supervision time changes little. This task split matches the scheduling model: retiming shortens an episode when raw cross-arm serialisation constrains its critical path; otherwise, it changes joint supervision without materially changing duration.

Because the policy predicts synchronised three-arm action chunks, the segment targets that share a chunk form its joint supervision. Co-window coverage measures this learner-facing change directly. The selected rollouts show the behavioural counterpart: the learned policies advance one stream while retaining tray stabilisation or fixture positioning where the task still requires it.

The robot trials evaluate the complete Raw-to-DATS intervention. The Gap audit shows that DATS changes segment placement relative to an order-preserving timeline with idle intervals removed, while the co-window analysis characterises the resulting joint targets. In the evaluated setting, DATS aligns pairwise-collected demonstrations with synchronous tri-manual execution.

\section{Conclusion}
We presented \emph{TriManPolicy}, which treats mode-switch staging as a property of the training timeline rather than a restriction that the deployed three-arm policy must inherit. By preserving demonstrated local behaviours while changing their temporal composition, DATS presents graph-permitted cross-arm progress within common training windows. Across six tasks, the evaluated DATS policies are faster on successful executions and record equal or higher observed success counts in every task. These results show that supervision timing matters in pairwise-collected tri-manual imitation.

A demonstration records constraints imposed by its collection interface as well as the task itself. When those constraints do not apply at deployment, graph-permitted retiming offers a way to keep collection-induced staging from being inherited by the learned policy.

\bibliographystyle{IEEEtran}
\bibliography{refs}

\end{document}